\documentclass[10pt, a4paper]{article}

\usepackage[final]{lrec2026}
\usepackage{color}
\usepackage{enumitem}
\usepackage{booktabs}
\usepackage{multirow}
\usepackage{tabularx}
\usepackage{array}
\usepackage{xcolor}
\usepackage{colortbl}
\usepackage{tikz}
\usetikzlibrary{positioning}

\title{\emph{Queering the Audits}: Community-Based Auditing of AI Harms to Queer Communities}

\name{Organizers of QueerInAI$^{\ast}$, A Pranav$^{\ast\dagger}$, Alissa A. Valentine$^{\ast\ddagger}$, Alex Markham$^{\ast\ddagger}$, \\
      {\bf \large Beckett LeClair$^{\ast\S}$, Tereza Blazkova$^{\diamond\ddagger}$, Ekaterina Kornilitsina$^{\diamond\P}$,} \\
      {\bf \large Sofie H. Bruun$^{\diamond\|}$, Gerasimos Spanakis$^{\diamond\clubsuit}$, and Anne Lauscher$^{\diamond\dagger}$}}

\address{$^{\dagger}$University of Hamburg \hspace{1em}
         $^{\ddagger}$University of Copenhagen \hspace{1em}
         $^{\S}$5Rights Foundation \\
         $^{\P}$Independent Researcher \hspace{1em}
         $^{\|}$Alexandra Instituttet \hspace{1em}
         $^{\clubsuit}$Maastricht University \\
         \thanks{$^{\ast}$Equal contribution.} \thanks{$^{\diamond}$Equal advising.}}

\abstract{
AI systems embed majority-group defaults into training data, evaluation metrics, and category definitions, producing documented harms for queer communities including erasure, misclassification, and discrimination.
Standard technical audits often rely on aggregate measures and cannot detect harms that become visible only through the lived experience of affected communities.
We conducted a participatory auditing workshop at EurIPS 2025 where 16 queer community members audited four case studies using the 4Cs harm taxonomy (\textbf{\textsc{Content}}, \textbf{\textsc{Conduct}}, \textbf{\textsc{Contact}}, \textbf{\textsc{Contract}}) applied across the AI lifecycle.
Participants used structured worksheets and plenary synthesis to classify harms and trace them to their origins in the development pipeline.
Across all four cases, participants traced harms to problem definition and data collection, and they identified contractual structures that extract value from vulnerable populations while providing minimal recourse.
These findings illustrate that community-informed auditing surfaces identity-specific harms that aggregate evaluation methods risk overlooking.
\\ \newline  }

\begin{document}

\maketitleabstract

\section{Introduction}

Identity-aware AI development requires recognizing that standard machine learning pipelines embed majority-group defaults into training data, category definitions, and evaluation metrics \citep{mundt2025cakeintelligencegetsbake}.
For queer communities, these defaults produce documented harms: training data under-represents queer experiences, producing systems that default to heteronormative and cisnormative assumptions \citep{taylor2024cruising, felkner2023winoqueer}; rigid demographic categories fail to represent fluid identities \citep{keyes2018misgendering, qinai}; and content moderation and recommendation systems disproportionately restrict queer expression \citep{southerton2021restricted,Mayworm2024}.

These harms persist because standard evaluation methods and technical audits often rely on aggregate measures and miss harms that become visible only through the lived experience of affected communities \citep{birhane_power_2022}.
Participatory auditing addresses this gap by including affected communities directly in the evaluation process \citep{hartmann_addressing_2025, delgado_participatory_2023}.

\begin{figure}
    \centering
    \includegraphics[width=\linewidth]{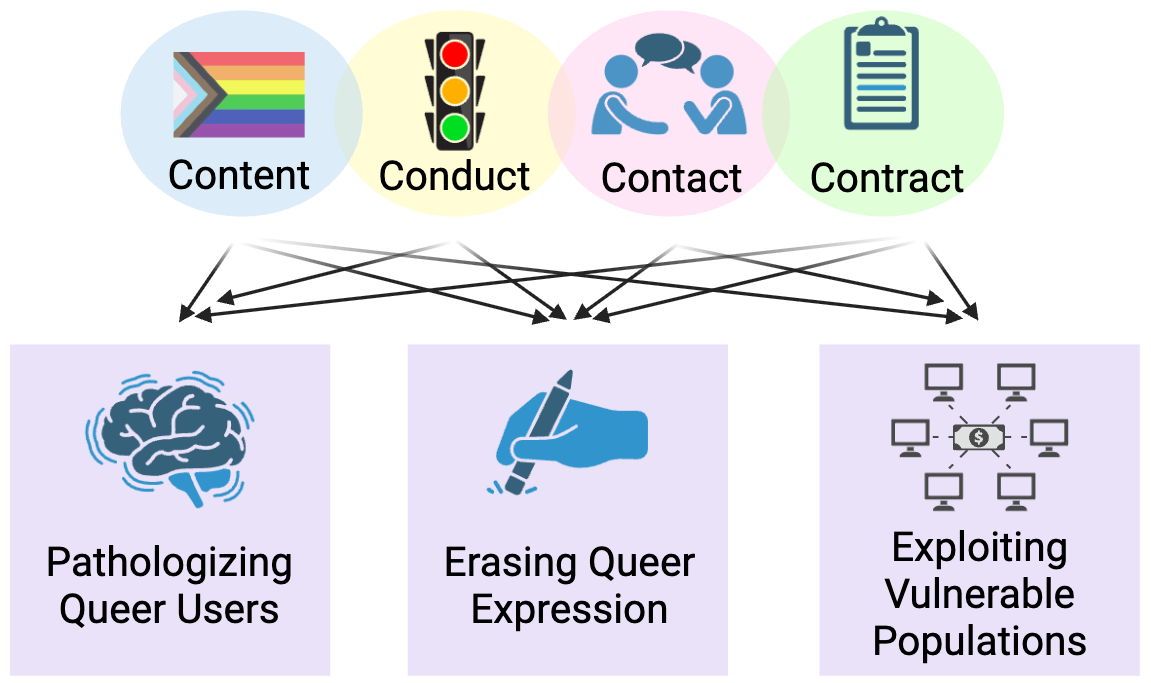}
    \caption{Overview of community-based AI auditing using the 4Cs (Content, Conduct, Contact, Contract) framework.}
    \label{fig:4cs_overview}
\end{figure}

This paper asks: \textit{What identity-specific harms do AI systems produce for queer communities, and how does structured community auditing surface them?}
We address this through a participatory auditing session conducted at the Queer in AI workshop, EurIPS 2025.\footnote{\url{www.queerinai.com/eurips-2025}}
16 queer community members examined four scenarios (mental health chatbots, ad-supported chatbots, content moderation, and data annotation labor) using the 4Cs harm taxonomy  (\textbf{\textsc{Content}}, \textbf{\textsc{Conduct}}, \textbf{\textsc{Contact}}, \textbf{\textsc{Contract}}) \citep{4Cs} applied across the AI lifecycle (Figure~\ref{fig:4cs_overview}).
This paper contributes a practical demonstration of participatory auditing with participants who have technical AI backgrounds but no formal auditing training, and documents the identity-specific harms it surfaced across four cases.

\section{The 4Cs Framework} \label{sec:4c_framework}

Participatory auditing requires a harm taxonomy that non-expert participants can apply consistently across different AI systems.
The 4Cs framework, originally developed to classify risks to children in digital spaces \citep{4Cs}, organizes harm by the \textit{relationship} between system and user rather than by technical properties of the system itself.
This relational focus is well suited to participatory auditing, where community members assess harms from their inherent lived experience rather than through less accessible technical metrics \citep{birhane_power_2022}.
Unlike taxonomies organized around technical system properties \citep{weidinger2022taxonomy, shelby2023sociotechnical}, the 4Cs foreground the user's position relative to the system, making them accessible to participants without requiring knowledge of model internals.
We classify harms along four dimensions:
\begin{enumerate}[noitemsep]
    \item \textbf{\textsc{Content}} harms concern what AI systems produce, label, flag, or promote. For queer users, these include chatbot responses that pathologize queer identities, and moderation systems that suppress queer expression.
    \item \textbf{\textsc{Conduct}} harms concern what behaviors AI systems enable or normalize among users and operators. These include leveraging intimate user disclosures for micro-targeted advertising and facilitating creation of non-consensual deepfakes.
    \item \textbf{\textsc{Contact}} harms concern the connections AI systems mediate or sever. These include recommendation algorithms that out queer users without consent and chatbot interactions that displace rather than supplement professional support.
    \item \textbf{\textsc{Contract}} harms concern the terms governing user engagement. These include opaque privacy policies for mental health apps and exploitative labor conditions for data annotators.
\end{enumerate}
These categories can overlap; we classify harms by their primary mechanism and note cross-cutting patterns in Section~\ref{sec:discussion}.

The decision was taken to attempt this activity through the lens of the 4Cs framework for three main reasons. Firstly, the categories of harm are not only specific to minors and can manifest for any group of users, be they members of a vulnerable group or otherwise. This made it a convenient choice of framework for considering all the potential avenues through which harms may manifest. Secondly, though the context specifics vary significantly, both children and queer communities have complex histories of being denied agency, autonomy and dignity in social settings - this made the framework seem a fitting choice. And finally, it is important to recognise there is an overlap between the queer community and the global community of children. Some of our findings can apply to queer youth as well as older people, and in some cases disproportionately so.

\section{Methods} \label{sec:methods}

Building on the auditing methodology from \citet{queerinai2023boundbountycollaborativelyshaping}, we conducted a 90-minute participatory auditing workshop at the Queer in AI workshop at EurIPS 2025.
Participants (N=16, across four groups) were queer community members attending the conference, with backgrounds in natural language processing, computer vision, social computing, and AI policy-making.
While participants had technical AI expertise, none had professional experience developing or deploying the specific system types under audit, nor formal training in auditing methodology.
Participants thus combined domain-adjacent technical knowledge with lived experience of queer-specific harms.
Each group was assigned one of four case study scenarios by the workshop organizers, matching groups to scenarios based on participants' self-reported expertise (e.g., participants with NLP backgrounds were assigned the chatbot scenarios, those with computer vision backgrounds the content moderation scenario).
All participants provided informed consent for their responses to be used in research.

\begin{figure}
    \centering
    \includegraphics[width=\linewidth]{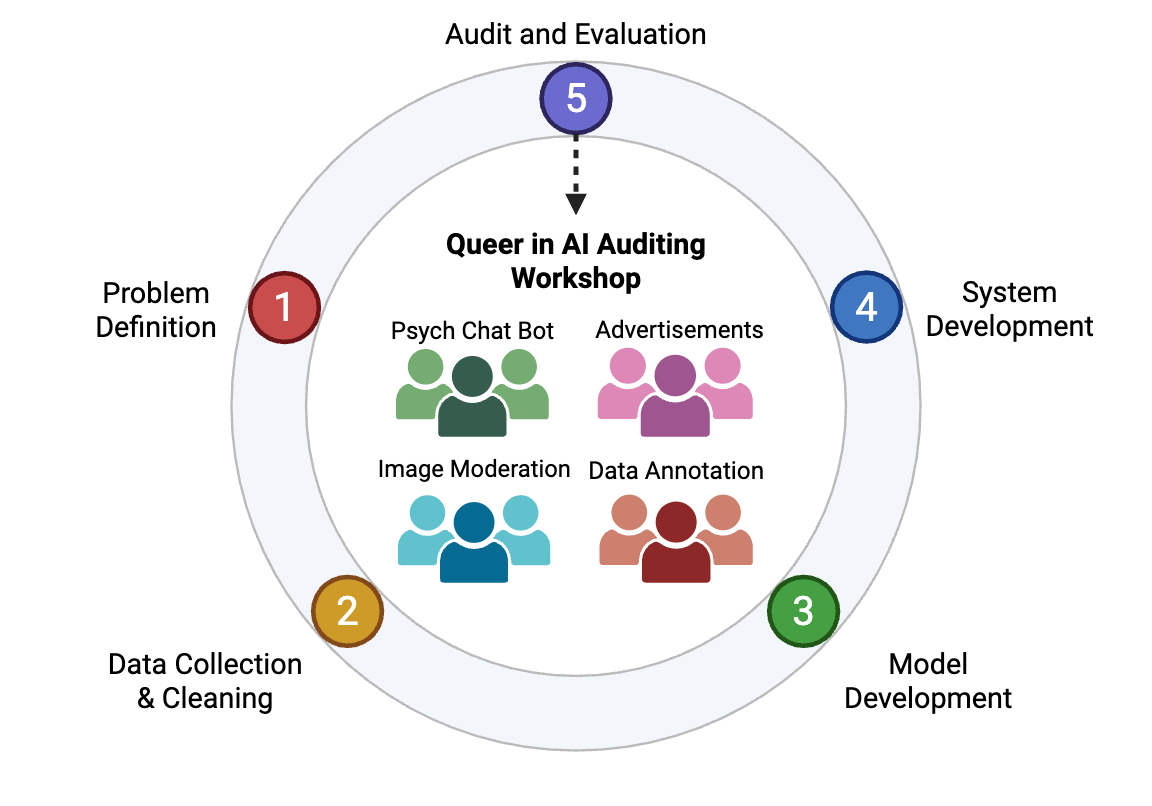}
    \caption{The workshop positions community auditing at the fifth stage of the AI lifecycle. The audit activity itself occurs at this stage, but participants examine decisions made at all prior stages: Problem Definition, Data Collection, Model Development, and Deployment.}
    \label{fig:lifecycle}
\end{figure}

\subsection{Analytical Framework}

Groups classified harms using the 4Cs taxonomy (Section~\ref{sec:4c_framework}) applied across a five-stage AI lifecycle (Figure~\ref{fig:lifecycle}): Problem Definition, Data Collection, Model Development, Deployment, and Evaluation.
This follows ecosystem auditing methods that trace harms across the full development pipeline rather than evaluating outputs alone \citep{Ojewale_2025}.
Each lifecycle stage was paired with guiding questions adapted from \citet{raji2022outsideroversightdesigningparty}, such as ``Who decided this problem needs solving?'' (Problem Definition), ``Whose data is included?'' (Data Collection), and ``Who is evaluating and are affected communities involved?'' (Evaluation).

\subsection{Case Studies}
The workshop organizers at Queer in AI collaboratively developed four case study scenarios, selecting systems along the axes of vision, natural language processing (NLP), data annotation, and human-computer interaction (HCI) to cover distinct modalities of AI harm relevant to queer communities.
Each scenario represents an AI system with documented relevance to queer communities.
The first three examine harms to users of AI systems; the fourth examines harms to workers who build them:

\begin{enumerate}[noitemsep,nolistsep]
    \item A mental health chatbot trained on therapist-patient transcripts, deployed for queer youth.
    \item A free AI chatbot monetizing through personalized ads based on disclosed identity, relationships, and mental health struggles.
    \item A content moderation system that classifies and removes user-generated content, with documented disparities affecting queer creators and bodies.
    \item Data annotation workers labeling toxic and hateful content to train moderation models, exposed to psychological harm for low pay.
\end{enumerate}
These scenarios describe system types rather than specific products, allowing participants to draw on their collective knowledge of real-world examples while avoiding the constraints of auditing a single deployment.

\subsection{Procedure}

Each group received a scenario description and a structured worksheet (Appendix \ref{app:worksheet}).\footnote{\url{github.com/valentinealissa/queerinai}}
After an introductory presentation on the 4Cs framework and AI lifecycle, groups spent 30 minutes classifying harms using the 4Cs at each lifecycle stage and proposing interventions, then distilled their top findings and prepared a plenary summary (10 minutes).
The 30-minute analysis window constrained depth; however, the goal was to surface harms that participants recognized from lived experience rather than to conduct exhaustive technical evaluation.
In the plenary, each group reported their most significant findings, the lifecycle stages where intervention could help, and one question they would pose to developers.
A facilitated cross-case discussion followed, structured around three prompts: which harm categories recurred, whether harms clustered at particular lifecycle stages, and what patterns emerged across systems.

\subsection{Data Collection and Analysis}

Participants submitted responses via structured online worksheets capturing identified harms, lifecycle mappings, intervention proposals, and auditing questions.
Each of the four groups submitted a complete response for their respective case studies.
Facilitators documented the plenary discussion.
We organized responses by case study and coded each harm using the 4Cs taxonomy, then compared findings across case studies to identify the cross-cutting patterns reported in Section~\ref{sec:discussion}.

\subsection{Limitations} \label{sec:limitations}

Firstly, the workshop involved a small number of participants (N=16) recruited from a single conference. Therefore, conference attendees are not fully representative of queer communities broadly. In addition, we did not survey the workshop attendees for their demographic information to be able to make conclusions about the representation of intersectional identities.
Participants had technical AI backgrounds (NLP, computer vision, AI policy), and groups were assigned to scenarios matching their expertise. The findings therefore reflect domain-adjacent professionals applying lived experience, not lay community members encountering these systems for the first time. Future work should test whether participants without technical backgrounds surface comparable harms.

Another limitation is the structure and duration of the auditing session. The 30-minute analysis window limited the depth of each group's audit.
The structured worksheets and guiding questions may have directed participants toward similar findings across groups, contributing to the convergent patterns reported in Section~\ref{sec:discussion}. This convergence may therefore reflect the shared analytical structure rather than independently surfaced patterns alone.

Participants audited system types rather than specific deployed products, which increases generalizability across system categories but limits the specificity and actionability of findings.
Although we found inspiration from existing AI systems to develop the four case studies in this audit, we do not provide participants with specific real-world examples. For example, Case Study 3 mimics real world scenarios, but is not an audit of the existing content moderation policies of Meta. We believe this approach allowed us to build upon collective experiences with real-world systems, without enforcing the audit of one single instance of an AI system.

We did not compare our participatory approach to a non-participatory audit of the same systems, so we cannot empirically demonstrate that community-informed auditing surfaces harms that technical audits miss, only that it surfaces harms consistent with documented patterns in the literature.
We therefore present this work as a proof-of-concept for structured participatory auditing with queer communities rather than as definitive evidence of the method's superiority over other approaches.

\section{Results} \label{sec:results}

\subsection{Case Study 1: Therapist Chatbots for Queer Youth} \label{sec:case_1}
Queer youth face significant barriers to accessing affirming mental healthcare, and among LLM users who self-report mental health issues, 49\% report using these systems for support \citep{sentio2025survey}.
Four workshop participants audited this scenario: a chatbot trained on therapist-patient transcripts and marketed to queer youth seeking affirming care.

\noindent\textbf{\textsc{Content}.}
Participants raised concerns that training data composition shapes therapeutic outputs.
The clinical workforce from which training transcripts derive lacks diversity \citep{lin2018diverse}, and historical pathologization of queer identities persists in clinical literature \citep{drescher2015out}.
Because therapist demographics shape therapeutic approaches and language, non-diverse workforces produce training corpora that reflect a narrow range of clinical perspectives.
Participants argued that training corpora drawn from this context risk encoding the same biases that drove queer users away from traditional therapy and questioned whether developers had audited for pathologizing content.

\noindent\textbf{\textsc{Conduct}.}
Participants raised concerns about deploying unvalidated systems to populations in crisis.
Queer youth experience depression at significantly higher rates than heterosexual peers and report elevated rates of suicidal ideation \citep{ma2024evaluating}.
Yet the systems marketed to them lack clinical validation: a systematic review of 160 AI chatbot studies found only 16\% underwent clinical efficacy testing \citep{hua2025charting}.
An evaluation of 29 mental health chatbots found that none met the criteria for adequate suicide risk response \citep{pichowicz2025performance}.
Participants observed that this validation gap is compounded by user vulnerability: people in crisis are more likely to trust an LLM that appears empathetic, following advice they would reject from a visibly unqualified human.

\noindent\textbf{\textsc{Contact}.}
Participants observed that chatbot interactions displace rather than supplement professional care.
In text-based exchanges, users rate LLM-generated responses as more empathetic than those from licensed therapists (in a social media forum context, not clinical settings; \citealp{ayers2023comparing, hasan2025llmtherapists}), and one in four Americans report preferring AI chatbots over therapists \citep{iftikhar2024therapy}.
Participants traced this preference to a structural mismatch: effective therapy requires confrontation of harmful thought patterns, yet LLMs optimized for user satisfaction produce sycophantic responses \citep{moore2025expressing, malmqvist2024sycophancy}.
These responses reinforce rather than challenge maladaptive cognition.
Queer youth, who already face barriers to affirming care, may gravitate toward validation over therapeutic challenge, reducing the likelihood they seek the professional support they need.
Participants acknowledged an important counterpoint: for queer users in unsupportive environments, where affirming therapists are unavailable or consulting a mental health professional carries safety risks, chatbot access may outweigh documented harms. 

\noindent\textbf{\textsc{Contract}.}
Participants identified exploitative terms governing vulnerable users' engagement with these systems.
Empirical analysis found 74\% of mental health applications scored ``Critical Risk'' for privacy, with policies requiring college-level education to comprehend \citep{iwaya2023privacy}. Youth are particularly vulnerable to exploitative privacy terms given documented age-related differences in privacy comprehension and risk assessment \citep{livingstone2011risks}.
Participants noted that mental health chatbots often fall outside existing health data protection frameworks such as HIPAA \citep{marks2023ai}, leaving users without meaningful informed consent or recourse.
Participants concluded that without proper auditing, deployment decisions go unchecked and vulnerable populations bear the consequences of systems never assessed for the harms they produce.

\subsection{Case Study 2: Ad-Supported Chatbots} \label{sec:case_2}

A free AI chatbot monetizes by serving personalized ads based on intimate conversational disclosures, including identity, relationships, mental health struggles, and financial stress.
Four workshop participants audited this scenario for harms to queer users.

\noindent\textbf{\textsc{Content}.}
Participants identified that ad targeting based on intimate disclosures can surface content that directly contradicts user needs.
They gave the example of users who disclose gambling problems receiving casino promotions, or those discussing debt being steered toward predatory financial products.
For queer users, identity-based targeting compounds this risk: research documents that ad systems can surface conversion therapy services and content reinforcing stereotypes about queer people \citep{via2022conversion}, normalizing discrimination through repeated exposure.

\noindent\textbf{\textsc{Conduct}.}
Participants raised concerns about a fundamental conflict of interest: the system simultaneously serves as a source of support and as an advertising platform, yet users cannot distinguish when recommendations serve commercial incentives rather than care.
They observed that the system enables manipulative practices by leveraging disclosures shared under an assumption of confidentiality \citep{christopherson2007positive}, such as coming-out experiences or mental health crises, to craft targeted nudges toward consumption.
Participants noted that political and ideological advertisers can further exploit these profiles to micro-target users with tailored disinformation \citep{woolley2016automating}.

\noindent\textbf{\textsc{Contact}.}
Participants discussed how ad-driven curation shapes which communities and resources users encounter.
They raised concerns that queer users may be steered toward hostile or exploitative spaces while affirming communities and support resources are deprioritized, isolating users from healthier networks.
Participants noted that this dynamic is particularly consequential for users whose primary social outlets are online rather than offline \citep{boyd2014s}, for instance due to a lack of queer-supportive spaces in their immediate environment.

\noindent\textbf{\textsc{Contract}.}
Participants identified that users trade deeply personal data for access to a ``free'' service without meaningful choice.
Terms of service are written to obscure how disclosures will be used for profiling, and opting out may mean losing core features.
Participants discussed that the platform thus extracts significant value from users' disclosures \citep{bodle2016critical, zuboff2015big} while providing uncertain quality of care and minimal recourse, a dynamic that disproportionately affects marginalized users with fewer alternatives.

\subsection{Case Study 3: Content Moderation} \label{sec:case_3}

A content moderation model is trained to automatically flag and remove images classified as ``sexually explicit'' from social media.
Such moderation has expanded partly in response to policies like the 2008 PROTECT Our Children Act \citep{Thakor2023To,PROTECT}, but these practices sometimes disproportionately target queer creators and bodies \citep{Haimson2021,Mayworm2024,Dias}.
Four workshop participants audited this scenario.

\noindent\textbf{\textsc{Content}.}
Participants raised concern that moderation systems encode cisnormative assumptions about acceptable bodies and expression, systematically associating queerness with sexual deviancy \citep{berro2024toward}.
They argued that when these biases are embedded in automated classifiers, queer bodies are flagged independent of context, leading to disproportionate removal of content from drag performers, transgender creators, and queer sex workers \citep{ungless-etal-2023-stereotypes}.

\noindent\textbf{\textsc{Conduct}.}
Participants identified problem definition as the most consequential stage: who defined ``sexually explicit,'' and whose norms does that definition encode?
Without public documentation of classification criteria, users have no way to determine whether definitions reflect broad community standards or encode the cultural biases of a narrow set of decision-makers.
Participants also observed that this opacity results in content creators learning to self-censor in anticipation of algorithmic removal, normalizing the exclusion of queer expression from public spaces.

\noindent\textbf{\textsc{Contact}.}
Participants observed that content moderation directly affects queer people exploring their identity and seeking community.
Social media offers spaces for queer people, especially queer youth, to find community and overcome offline marginalization driven by stigma or safety concerns \citep{miller2017my,hanckel2014finding}.
Moderation that fragments these spaces severs the conversations and connections they sustain, risking isolation for people whose primary community access is online.

\noindent\textbf{\textsc{Contract}.}
Participants noted that opaque moderation policies leave queer content creators unable to contest removal decisions or understand what triggered them \citep{suzor2019lawless}.
They called for post-deployment feedback mechanisms that allow community members to report incorrect flags and receive timely responses.
Participants identified a structural asymmetry: queer creators risk losing livelihoods while platforms benefit from the engagement their content generates.

\subsection{Case Study 4: Data Annotation for Moderation} \label{sec:case_4}

Data annotation is the human labor that underpins multiple stages of the AI lifecycle: annotators label training data, validate model outputs, and evaluate system performance.
In content moderation pipelines, annotators label content as ``toxic,'' ``hateful,'' or ``violent'' to train models what to flag, filter, or remove.
Unlike the other case studies, which examine harms to AI users, this case study examines harms to the workers who build AI systems.
Four workshop participants audited this scenario: a data annotation pipeline for content moderation, where workers are routinely exposed to psychologically harmful material under exploitative conditions.

\noindent\textbf{\textsc{Content}.}
Participants raised concerns that queer annotators face concentrated exposure to homophobic and transphobic hatred as routine labor, compounding the stigma-related stressors queer workers already experience \citep{meyer2003prejudice}.
They noted that even individually minor content accumulates into psychological harms including anxiety, depression, and PTSD \citep{steiger2021psychological,Cambridge}. 
Participants also identified how annotator biases flow downstream.
If annotators view queer bodies as more obscene than non-queer bodies, these judgments become embedded in model behavior, producing moderation disparities that restrict queer expression \citep{Dorn}. 

\noindent\textbf{\textsc{Conduct}.}
Participants observed that binary labeling frameworks (``harmful'' / ``not harmful'') force oversimplification of culturally situated content.
They gave examples such as drag performances and queer health discussions being labeled harmful by annotators unfamiliar with queer culture, while equivalent heterosexual content passes without scrutiny.
Participants emphasized that annotators who hold homophobic views, or who simply lack familiarity with queer culture, encode their biases into model behavior \citep{GLAAD,Dias}.
They advocated for multidimensional labeling schemes as an alternative to binary classifications.

\noindent\textbf{\textsc{Contact}.}
Participants discussed how non-disclosure agreements and stigma isolate annotation workers from professional support \citep{Perrigo2023}.
They highlighted that queer annotators in hostile regions cannot disclose the nature of their distress without risking personal safety, a dynamic consistent with concealment as a proximal minority stressor \citep{meyer2003prejudice}.
Participants further noted that biased annotation has consequences beyond the annotator: if queer content is mislabeled as toxic, moderation systems remove it, and queer users lose the shared online spaces where they find community \citep{miller2017my,hanckel2014finding}.

\noindent\textbf{\textsc{Contract}.}
Participants identified exploitative labor conditions, noting that investigative reporting documents Kenyan annotation workers earning less than \$2 per hour with inadequate psychological support \citep{Perrigo2023}.
They observed that layers of subcontracting distance AI companies from responsibility for these conditions \citep{posada2022coloniality}.
Participants raised particular concern for queer workers in criminalizing jurisdictions, who cannot organize, report identity-specific harms, or seek legal recourse without exposing themselves to prosecution \citep{ILGAWorld2024}.

\section{Discussion} \label{sec:discussion}

Following the case study audits, each group presented their findings in a plenary session.
A facilitated discussion then prompted participants to compare findings across all four systems.
Three cross-cutting themes emerged.

\textbf{\textit{Harms mainly cluster during problem definition and data collection.}}
Participants traced harms to problem definition and data collection in all four case studies.
The mental health chatbot inherited pathologizing norms from clinical training data produced by a non-diverse workforce (Section~\ref{sec:case_1}).
The moderation system encoded culturally specific definitions of ``sexually explicit'' before any model was trained (Section~\ref{sec:case_3}).
The ad-supported chatbot treated disclosures as targeting material because its revenue model required it (Section~\ref{sec:case_2}).
Binary annotation frameworks reflected annotator biases that then flowed into model behavior (Section~\ref{sec:case_4}).
This finding echoes prior analyses of harm propagation across ML pipelines \citep{suresh2021framework}, but our workshop illustrates how community participants arrive at the same conclusion through lived experience rather than technical analysis.
Participants concluded that auditing model outputs alone misses root causes; auditing must also examine the assumptions, data, and incentive structures that shaped the system before any code was written.
This aligns with what \citet{birhane2024ai} term ``ecosystem audits'': investigations that go beyond datasets, models, and products to examine the communities and sociotechnical environments defining an AI system's operation.

\textbf{\textit{Contractual structures extract value from vulnerable populations.}}
Across case studies, participants observed a recurring asymmetry: the populations most exposed to harm had the least ability to understand, negotiate, or challenge the terms governing their interactions with these systems.
Users disclosed mental health struggles under opaque terms of service (Sections~\ref{sec:case_1},~\ref{sec:case_2}).
Content moderation systems erased queer expression while platforms benefited from the engagement that queer creators generated (Section~\ref{sec:case_3}).
Annotation workers endured psychological harm for low pay while their labor built the systems that moderate queer expression (Section~\ref{sec:case_4}).
In each case, a lack of transparency about system development and deployment prevented users and workers from contesting the terms they faced.
These patterns suggest that contract harms follow from business models that treat vulnerable populations' data and labor as extractable inputs rather than as interests to protect.

\textbf{\textit{Existing governance frameworks do not account for identity-specific harms.}}
The EU AI Act classifies systems by risk tier \citep{EUAIAct2024}, and international frameworks emphasize transparency and oversight \citep{UNESCO2021,OECD2019}.
Under these frameworks, the risk classification of the four systems participants audited is uncertain: some, such as the mental health chatbot, may fall under high-risk categories depending on regulatory interpretation, while others operate under platform self-regulation or fall outside existing mandates entirely.
Yet all produced harms that participants considered significant.
Governance frameworks organized around system-level risk categories may not adequately capture harms that emerge at the intersection of system design and user identity.
This gap reflects documented failures to translate AI ethics principles into mechanisms that protect specific populations \citep{Mittelstadt2019,Hagendorff2020}.

\section{Recommendations} \label{sec:recommendations}

Our findings suggest three recommendations for identity-aware AI development.
Although these reflect queer community concerns, they apply to any population underrepresented in AI development.

\textbf{\textit{Include affected communities throughout the lifecycle.}}
The most consequential decisions occur at problem definition and data collection, yet affected communities are typically consulted only after deployment \citep{sloane2022participation}.
In our workshop, participants identified training data biases in the mental health chatbot that would require lived experience to recognize, because distinguishing pathologizing clinical norms from affirming ones demands familiarity with their effects (Section~\ref{sec:case_1}).
Community involvement should extend beyond post-deployment feedback to inform problem definition, evaluation criteria, and ongoing governance.

\textbf{\textit{Recognize identity throughout the pipeline.}}
Annotator biases that participants identified in Case Study 4 can flow into the moderation disparities documented in Case Study 3, illustrating how identity-blind assumptions at one stage can produce identity-specific harms at another.
Training data composition determines what the system treats as normal (Section~\ref{sec:case_1}).
Operationally, this means auditing annotator pools for demographic and cultural diversity, evaluating the impacts of annotator positionality, and documenting assumptions about identity at each lifecycle stage \citep{riossialer2026structure}.

\textbf{\textit{Audit problem definitions, not just outputs.}}
Prior work has argued that the question ``should we build this?'' should precede ``how do we build this?'' \citep{selbst2019fairness, green2022escaping}.
Our workshop provides empirical grounding for this recommendation: in all four case studies, participants traced the most significant harms to choices made before model development began.
For applications affecting vulnerable populations, regulations should be implemented to ensure that accountability mechanisms and ethical review must be established before production, not retroactively after harm has occurred.

\section{Conclusion} \label{sec:conclusion}

We demonstrate a practical framework for participatory auditing: community members without auditing expertise, given a structured harm taxonomy and guided questions, identified concrete harms grounded in lived experience.
\textbf{Audits are one mechanism among many for achieving accountability, but they are most effective when they include affected communities}, extend beyond technical evaluation, and connect findings to concrete demands for change.
Future work should test this approach at larger scale, apply it to specific deployed systems rather than hypothetical scenarios, and investigate how community audit findings can be integrated into existing governance processes.

\section{Bibliographical References}\label{sec:reference}
\bibliography{main}
\bibliographystyle{lrec2026-natbib}

\appendix
 
\section{Auditing Worksheet}
\label{app:worksheet}
 
The following worksheet was distributed to participants at the Queer in AI auditing workshop at EurIPS 2025.
Participants completed it in groups of two to three over a 30-minute structured analysis window.
 
\bigskip
 
\noindent\textbf{Harm taxonomy: The 4Cs}
 
\medskip
 
\noindent There are many ways of classifying digital harms. The 4Cs taxonomy organizes harm by the
relationship between system and user. The categories can overlap; a fifth category, \textit{Cross-Cutting},
captures harms that span multiple dimensions.
 
\medskip
 
\noindent
\begin{tikzpicture}[
  every node/.style={align=center, font=\small},
  box/.style={
    draw, rounded corners=6pt,
    text width=0.42\linewidth,
    minimum height=3.4cm,
    inner sep=8pt
  },
  pink/.style={box, fill=red!10, draw=red!30},
  yellow/.style={box, fill=yellow!20, draw=yellow!50},
  green/.style={box, fill=green!12, draw=green!40},
  blue/.style={box, fill=blue!10, draw=blue!30}
]
 
\node[pink] (content) at (0,0) {
  \textbf{Content}\\[4pt]
  Things made or shared with help from AI systems\\[6pt]
  \textit{Examples:} Generated text or images perpetuating harmful
  stereotypes; anti-queer content promoted by recommendation algorithms
};
 
\node[yellow, right=0.04\linewidth of content] (conduct) {
  \textbf{Conduct}\\[4pt]
  How people behave using AI\\[6pt]
  \textit{Examples:} Mass-producing harmful disinformation;
  creating non-consensual deepfakes
};
 
\node[green, below=0.3cm of content] (contact) {
  \textbf{Contact}\\[4pt]
  Connections facilitated by AI\\[6pt]
  \textit{Examples:} Algorithmic friend recommendations that suggest
  anti-queer connections; outing users to known contacts without consent
};
 
\node[blue, below=0.3cm of conduct] (contract) {
  \textbf{Contract}\\[4pt]
  Terms users are subjected to\\[6pt]
  \textit{Examples:} Theft of queer creators' intellectual property for
  commercial gain; AI-powered behavior monitoring for ad revenue
};
 
\end{tikzpicture}
 
\bigskip
 
\noindent\textbf{Questions to ask at each lifecycle stage}
 
\begin{enumerate}[noitemsep, leftmargin=*, label=\arabic{enumi}.]
  \item \textbf{Problem definition:} Who decided this problem needs solving? Whose needs were centered?
  \item \textbf{Data collection and cleaning:} Whose data is included? Whose is missing? Was consent meaningful?
  \item \textbf{Model development:} What assumptions are encoded? How are edge cases handled?
  \item \textbf{System deployment:} Who has access? Who is excluded? What recourse exists?
  \item \textbf{Audit and evaluation:} Who is evaluating? What metrics matter? Are affected communities involved?
\end{enumerate}
 
\bigskip
 
\noindent\textbf{Scenarios (one assigned per group)}
 
\begin{itemize}[noitemsep, leftmargin=*]
  \item \textbf{Language:} A mental health chatbot trained on therapist-patient transcripts is rolled out to reduce psychiatric wait times, including for queer youth.
  \item \textbf{Vision:} A content moderation model automatically flags and removes images as ``sexually explicit,'' but disproportionately targets queer creators and bodies.
  \item \textbf{Data privacy:} A free AI chatbot monetizes by serving personalized ads based on what users disclose in conversation, including their identity, relationships, and mental health struggles.
  \item \textbf{Data collection:} Workers are hired to label toxic, hateful, and violent content so the model learns what to reject, exposing them to psychological harm for low pay.
\end{itemize}
 
\bigskip
 
\noindent\textbf{Worksheet questions}
 
\begin{enumerate}[noitemsep, leftmargin=*, label=\arabic{enumi}.]
  \item What harms could occur? (Use the 4Cs)
  \item Where in the lifecycle could these harms be introduced?
  \item Where could interventions have prevented them?
  \item What questions would you ask if you were auditing this system?
  \item What are your top 2--3 findings to share during plenary?
\end{enumerate}

\end{document}